\documentclass[preprints,article,accept,moreauthors,pdftex]{Definitions/mdpi} 

\firstpage{1} 
\pubvolume{xx}
\issuenum{1}
\articlenumber{5}
\pubyear{2019}
\copyrightyear{2019}
\history{Received: date; Accepted: date; Published: date}

\pdfoutput=1

\usepackage{subcaption}
\usepackage{xcolor}
\usepackage{siunitx}

\Title{Panoptic Instance Segmentation on Pigs}

\Author{Johannes Br\"unger $^{1,*}$\orcidA{}, Maria Gentz $^{2}$\orcidB{}, Imke Traulsen $^{2}$\orcidC{} and Reinhard Koch $^{1}$\orcidD{}}

\AuthorNames{Johannes Br\"unger, Maria Gentz, Imke Traulsen and Reinhard Koch}

\address{%
$^{1}$ \quad Department of Computer Science, Kiel University\\
$^{2}$ \quad Livestock Systems, Department of Pig Sciences, Georg-August-University G\"ottingen}

\corres{Correspondence: jobr@informatik.uni-kiel.de}

\abstract{The behavioural research of pigs can be greatly simplified if automatic recognition systems are used. Especially systems based on computer vision have the advantage that they allow an evaluation without affecting the normal behaviour of the animals. In recent years, methods based on deep learning have been introduced and have shown pleasingly good results. Especially object and keypoint detectors have been used to detect the individual animals. Despite good results, bounding boxes and sparse keypoints do not trace the contours of the animals, resulting in a lot of information being lost. Therefore this work follows the relatively new definition of a \emph{panoptic segmentation} and aims at the pixel accurate segmentation of the individual pigs. For this a framework of a neural network for semantic segmentation, different network heads and postprocessing methods is presented. With the resulting instance segmentation masks further information like the size or weight of the animals could be estimated. The method is tested on a specially created data set with 1000 hand-labeled images and achieves detection rates of around 95\% (F1 Score) despite disturbances such as occlusions and dirty lenses.}


\begin{document}


\section{Introduction}
There are many studies that show that the health and welfare of pigs in factory farming can be inferred from their behaviour. It is therefore extremely important to observe the behaviour of the animals in order to be able to intervene quickly if necessary. A good overview of the studies, the indicators found and the possibility of automated monitoring is provided by \cite{matthews_early_2016}. Similarly, there are studies in which the various environmental factors (housing, litter, enrichment) are examined and how these factors affect behaviour \cite{van_de_weerd_review_2009, veit_influence_2016, nasirahmadi_using_2017}.

Observing the behaviour of the animals over long periods of time cannot be done manually, so automated and sensor-based systems are usually used. Classical ear tags or collars can be located in their position, but have the disadvantage that the transmitter cannot provide information about the orientation of the remaining parts of the animal's body. In addition, the sensor must be purchased and maintained for each individual animal. This is why computer vision is increasingly used, where the entire barn with all animals can be monitored with a few cameras. An overview of different applications with computer vision in the pig industry can be found in \cite{nasirahmadi_implementation_2017}.

Based on 2D or 3D images, the position of the individual animals and their movements can be detected. From the positions alone, a lot of information can be extracted. By means of defined areas, the position can be used to identify e.g. food or water intake \cite{kashiha_automatic_2013}. Furthermore, interactions and aggression between the animals can be detected if they touch each other in certain ways (mounting, chasing) \cite{viazzi_image_2014, lee_automatic_2016, nasirahmadi_automatic_2016}. The behaviour of the entire group can also be evaluated. Certain patterns when lying down can reveal certain information about the temperature in the barn \cite{nasirahmadi_using_2015}. Or the changes in positions over time can be converted into an activity index \cite{ott_automated_2014} or locomotion analysis \cite{kashiha_automatic_2014}.

\begin{figure}[t]
\centering
\begin{subfigure}[t]{0.246\textwidth}
\includegraphics[width=\textwidth]{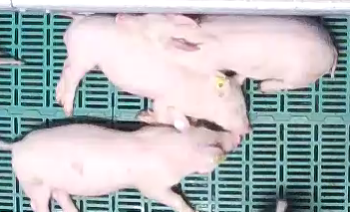}
\caption{\label{fig:intro_ellipses_bb_keypoints_a}Original image part}
\end{subfigure}
\begin{subfigure}[t]{0.246\textwidth}
\includegraphics[width=\textwidth]{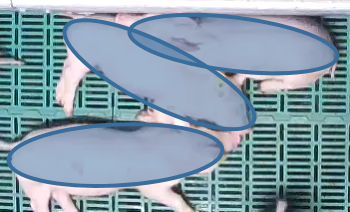}
\caption{\label{fig:intro_ellipses_bb_keypoints_b}Ellipses}
\end{subfigure}
\begin{subfigure}[t]{0.246\textwidth}
\includegraphics[width=\textwidth]{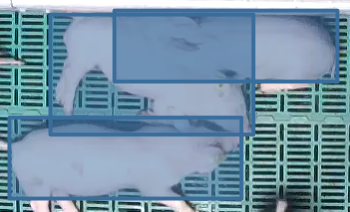}
\caption{\label{fig:intro_ellipses_bb_keypoints_c}Bounding boxes}
\end{subfigure}
\begin{subfigure}[t]{0.246\textwidth}
\includegraphics[width=\textwidth]{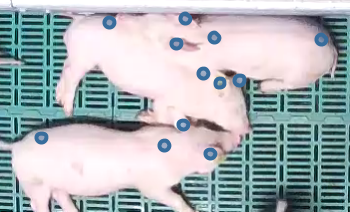}
\caption{\label{fig:intro_ellipses_bb_keypoints_d}Keypoints}
\end{subfigure}
\caption{\label{fig:intro_ellipses_bb_keypoints}Visualization of different types of detection. The proposed ellipses (b) provide more information about the pigs (like a weight-approximation) than the classic bounding boxes (large overlap) (c) or keypoints (d) where the affiliation to the individual animals has to be resolved afterwards.}
\end{figure}

Even though camera recording has many advantages due to its low-cost operation and non-intensiveness, the task of detecting animals reliably, even in poor lighting conditions and with contamination, is difficult. Previous work used classical image processing such as contrast enhancement and binary segmentation using thresholds or difference images to separate the animals from the background \cite{mcfarlane_segmentation_1995, shao_real-time_2008, kashiha_automatic_2013, nasirahmadi_automatic_2016}. Later, the advantages of more sophisticated detection methods based on learned features or optimization procedures were presented \cite{mittek_tracking_2018, brunger_model-based_2018}.
With the recent discoveries in the field of deep learning, the detection of pigs with neural networks has also been addressed. Either the established object-detector networks were applied directly to the pigs, or the detections found were post-processed to visually separate touching pigs \cite{ju_kinect-based_2018, zhang_automatic_2018, nasirahmadi_deep_2019}. Although the detection rate with these object detection methods is very good, the resulting bounding boxes are suboptimal, because depending on the orientation of the animal, the bounding box may contain large areas of background or even parts of other animals (see Figure \ref{fig:intro_ellipses_bb_keypoints}). Therefore, Psota et al. \cite{psota_multi-pig_2019} proposed a method that avoids the use of bounding boxes and tries to directly detect the exact pose of the animal with keypoints on specific body parts (e.g. shoulder and back).

In this work we close the gap between the too large bounding boxes and the sparse keypoints and try to identify the animals' bodies down to pixel level. We believe that the exact body outlines can help to classify the animals' behaviour even better. The movement of individual animals can be depicted much better than with a bounding box and the body circumference resulting from the segmentation can also be used to draw conclusions about the size and weight of the animals.

The main contribution of this thesis is the presentation of a versatile framework for different segmentation tasks on pigs together with the corresponding metrics.

The remainder of this work is organized as follows. In Section \ref{sec:background} the basic concepts of object detection based on bounding boxes, pixel-level segmentation and key-points are listed. The proposed method is described in Section \ref{sec:proposed_method} followed by the evaluation in Section \ref{sec:experimental_results}. The findings are discussed and concluded in Sections \ref{sec:discussion} and \ref{sec:conclusions}.

\section{Background}
\label{sec:background}
In recent years, methods based on neural networks have gained enormous importance in the field of image processing. Based on the good results in classification tasks, soon adapted network architectures were shown, which can also be used for the detection of objects \cite{sermanet_overfeat_2014, girshick_rich_2014}. The current generation of detection networks uses a combination of region proposals (bounding boxes) and classification parts, which evaluate the proposed regions \cite{liu_ssd_2016, ren_faster_2016, redmon_yolov3_2018}. With \emph{Deepmask} \cite{pinheiro_learning_2015, pinheiro_learning_2016} and \emph{Mask-RCNN} \cite{he_mask_2018} even object detectors have been shown, which generate a pixel-level segmentation mask for each region found.
Although these detectors provide very good results, the generated region proposals have the problem that only one object can be found at each position. This limitation is usually irrelevant, because in a projective image each pixel is assigned to exactly one object anyway and two objects at the same position cannot be seen. However, if two elongated objects overlap orthogonally, the center of the objects may fall on the same pixel, which cannot be mapped by such region proposal network.
\begin{figure}[t]
\centering
\begin{subfigure}[t]{0.19\textwidth}
\includegraphics[width=\textwidth]{assets/IntroPlain.png}
\caption{Original image part}
\label{fig:intro_ellipses_segmentation_modes_a}
\end{subfigure}
\hspace{.01cm}
\begin{subfigure}[t]{0.19\textwidth}
\includegraphics[width=\textwidth]{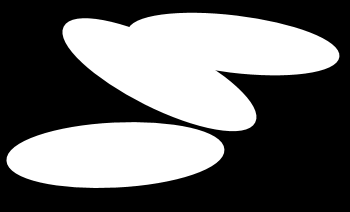}
\caption{Binary segmentation}
\label{fig:intro_ellipses_segmentation_modes_b}
\end{subfigure}
\hspace{.01cm}
\begin{subfigure}[t]{0.19\textwidth}
\includegraphics[width=\textwidth]{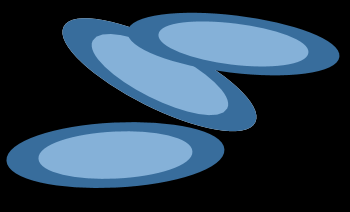}
\caption{Categorical segmentation}
\label{fig:intro_ellipses_segmentation_modes_c}
\end{subfigure}
\hspace{.01cm}
\begin{subfigure}[t]{0.19\textwidth}
\includegraphics[width=\textwidth]{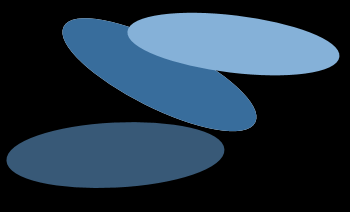}
\caption{Instance segmentation}
\label{fig:intro_ellipses_segmentation_modes_d}
\end{subfigure}
\hspace{.01cm}
\begin{subfigure}[t]{0.19\textwidth}
\includegraphics[width=\textwidth]{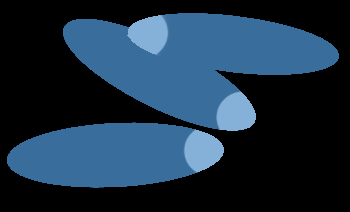}
\caption{Body part segmentation}
\label{fig:intro_ellipses_segmentation_modes_e}
\end{subfigure}
\caption{\label{fig:intro_ellipses_segmentation_modes}Visualization of the different experiments presented in this work. The binary segmentation distinguishes only between foreground and background (b). A categorical segmentation can be used to separate the individual animals (c) or to classify body parts (e). Or the network is trained to directly tell the affiliation of the pixels to the individual animals (d).}
\end{figure}
Another area in which neural networks are very successful is (semantic) segmentation, in which each pixel is assigned a class (pixelwise classification) \cite{long_fully_2015, ronneberger_u-net_2015, chen_rethinking_2017}. However, the classic semantic segmentation does not distinguish between individual objects, but only assigns a class to each pixel. In order to separate the individual objects, an instance segmentation must be performed.  For this purpose, the semantic segmentations are extended, for example, such that the output of the network is position-sensitive in order to identify the object boundaries \cite{li_fully_2017}. Another solution is to count and recognize the animals in a recursive way. For this purpose, one object after the other is segmented and stored until no more objects can be found \cite{romera-paredes_recurrent_2016, ren_end--end_2017}. Since the networks are designed to predict certain classes, the classes can also be chosen to help distinguish the instances. For example, Uhrig et al. \cite{uhrig_pixel-level_2016} use the classes to encode the direction to the center of the corresponding object for each pixel. Since the direction to the center of the object is naturally different at object boundaries, the individual instances can be separated.
To assign the pixels to individual instances an embedding can also be used. As described by De Brabandere et al. \cite{de_brabandere_semantic_2017}, a high-dimensional feature space is formed and for each pixel in the image the network predicts the position in space. Via a discriminative loss, pixels belonging to the same object are pushed together in the embedding space and pixel clusters of different objects are pushed apart. With a subsequent clustering operation the instances in the embedding can then be separated.

Another approach for the segmentation of individual instances is the detection of certain key points, which are then meaningfully combined into the individual instances using skeleton models \cite{papandreou_towards_2017, papandreou_personlab_2018, cao_realtime_2017}.

As described in the introduction, detection with bounding boxes and detection via key points has already been demonstrated on pigs. This work follows the relatively new definition of a \emph{panoptic segmentation} \cite{kirillov_panoptic_2019} and aims at the pixel accurate segmentation of the individual pigs.

\section{Proposed Method}
\label{sec:proposed_method}
The goal of the proposed method is a \emph{panoptic segmentation} \cite{kirillov_panoptic_2019} of all pigs in images of a downward-facing camera mounted above the pen. \emph{Panoptic segmentation} is defined as a combination of \emph{semantic segmentation} (assign a class label to each pixel) and \emph{instance segmentation} (detect and segment each object instance). So the semantic segmentation part differentiates between the two classes background and pig whereby the instance segmentation part is used to distinguish the individual pigs (see Figure \ref{fig:intro_ellipses_segmentation_modes_b} and \ref{fig:intro_ellipses_segmentation_modes_d}).

The proposed method for the \emph{panoptic segmentation} is an extension of classical semantic segmentation. Therefore, in this paper the complexity of segmentation is increased step by step resulting in four separate experiments. First a simple binary segmentation is tested (see Figure \ref{fig:intro_ellipses_segmentation_modes_b}). In the second experiment the individual animals are extracted from a semantic (or categorical) segmentation (see Figure \ref{fig:intro_ellipses_segmentation_modes_c}). The third experiment shows a pixel precise instance segmentation based on a combination of the binary segmentation and an pixel-embedding (see Figure \ref{fig:intro_ellipses_segmentation_modes_d}). And in the last experiment the embedding is combined with a body part segmentation (see Figure \ref{fig:intro_ellipses_segmentation_modes_e}) for an additional orientation recognition. 

\begin{figure}[t]
\includegraphics[width=\textwidth]{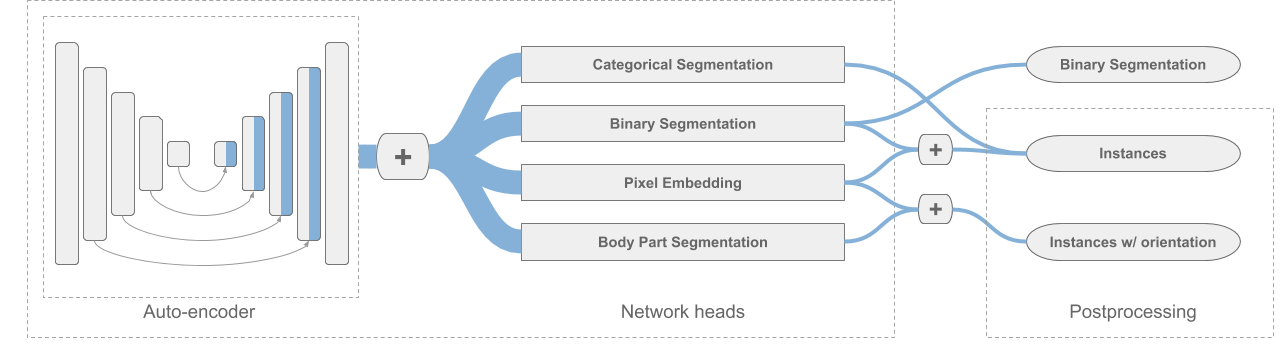}
\caption{\label{fig:framework_overview}Schematic representation of the proposed framework. The auto-encoder is an U-Net architecture (depiction adopted from \cite{yakubovskiy_segmentation_2019}). The individual stages consist of several blocks, each with several layers. Scaling down or up is done between the stages. Skip connections are used to combine the results of the encoder and decoder stages. The network is equipped with different heads for the different experiments. The output is processed afterwards to yield the desired results.}
\end{figure}

All experiments are based on the same network architecture. Only the last layers are adjusted to get the required output. This way, the presented framework can be easily adapted to each of the experiments. An overview of the framework is given in Figure \ref{fig:framework_overview}.

\subsection{Representation of the pigs}
\label{sec:representation_of_the_pigs}
In order to perform \emph{panoptical segmentation} instance by instance and as accurately as possible, manual annotation should contain such instance information with pixel accuracy. Since the choice of the annotation method always requires a trade-off between effort and accuracy, a pixel accurate annotation is preferable, but also very costly. In contrast, bounding boxes can be drawn quickly, but would contain large background areas in addition to the marked pig, especially if the pig is standing diagonally to the image axes (see Figure \ref{fig:intro_ellipses_bb_keypoints_c}). Based on existing work \cite{mcfarlane_segmentation_1995, kashiha_automatic_2013, nasirahmadi_using_2015}, ellipses were therefore chosen as annotations. They are also very easy to draw (due to the two main axes) and adequately reproduce the pigs' bodies on the images of a downward facing camera.  Except for small mistakes (e.g. when the animal turns its head to the side), the pixels belonging to the individual animals can thus be easily captured. By aligning the ellipse (first main axis), the orientation of the animals is also stored. If animals overlap, the order in which the pixels in the label image are drawn must correspond to the reversed order of the animals in the camera's visual axis. This ensures that the pixels of the animals on top overwrite the pixels of the covered animals (see Figure \ref{fig:intro_ellipses_segmentation_modes_c} and \ref{fig:intro_ellipses_segmentation_modes_d}). Since the area of the ellipses correlates approximately with the volume of the animals, the ellipses have the further advantage of allowing conclusions to be drawn about the volume respectively the weight of the animals.

\subsection{Network architecture}
\label{sec:network_architecture}
The typical network architecture for semantic segmentation consists of an encoder and a decoder part. The encoder transforms the input image into a low-dimensional representation, which the decoder then converts into the desired output representation (this combination is often also called auto-encoder). The encoder is structured similarly to a classification network whereas the decoder is a combination of layers symmetrical to the encoder, but with upsampling steps instead of the downsampling steps. To further improve the segmentation results, skip connections are often added. To make the information from the different resolution levels usable, these connections merge the intermediate results of the encoder with the corresponding upsampling levels in the decoder. Famous versions of such networks are for example U-Net \cite{ronneberger_u-net_2015} and LinkNet \cite{chaurasia_linknet_2017}. Another approach to use the information from the different downsampled layers from the encoder to obtain a dense segmentation in the original resolution is a feature pyramid network (FPN) \cite{lin_feature_2017}. Here the predictions are made at the different scales and merged afterwards.

In this work an U-Net was used as it gave the best results. In addition, its modular design allows the use of different classification architectures as encoders. Thus it is possible to benefit from the latest developments in this field. With ResNet34 \cite{he_deep_2015} and Inception-ResNet-v2 \cite{szegedy_inception-v4_2016} two established classification networks were used as encoder backbone. They both consist of single blocks that combine different convolution operations with a shortcut connection. With these shortcut connections, the optimizer does not have to learn the underlying mapping of the data, but simply a residual function \cite{he_deep_2015}. The blocks are organized in different stages and each stage is followed by a downscaling. The decoder part imitates the stages but uses an upscaling layer instead of downscaling. Via the skip connections the stages of the encoder are connected to the stages of the decoder where they are combined with the results of the encoder (see Auto-encoder in Figure \ref{fig:framework_overview}).

More details on the implemented architecture can be found in subsection \ref{sec:implementation_details}. In subsection \ref{sec:ablation_studies}, an ablation study evaluates additional backbones and hyper parameters as well as the FPN architecture.

\subsection{Binary segmentation}
A binary segmentation is the basis for many of the classical approaches to pig detection \cite{mcfarlane_segmentation_1995, shao_real-time_2008, kashiha_automatic_2013, nasirahmadi_using_2015}. At the same time, it is a comparably simple task for a neural network. Once solved, however, foreground segmentation can also be used to simplify more complex procedures, e.g. to apply them only to the important areas of the image (see subsection \ref{sec:instance_segmentation_with_discriminative_loss}).

For the binary segmentation, the network learns which pixels belong to the pigs and which to the background. So for each pixel $x_i$ it predicts a probability $p(x_i)$ with which the pixel belongs to a pig (with the corresponding opposite probability $(1 - p(x_i))$ the pixel belongs to the background). The training data consist of binary label images based on the manually annotated ellipses (see Figure \ref{fig:intro_ellipses_segmentation_modes_b}), where each pixel in the label image is a binary variable $y_i$, indicating whether the pixel belongs to the background (value 0) or to a pig (value 1).

The network is set up with the architecture described in section \ref{sec:network_architecture} but with only one output layer. The output has the same spatial dimension as the input, but with only one channel and a sigmoid activation function that generates the probability estimate for each pixel. The loss function is the cross-entropy loss:
\begin{equation}
L = -\frac{1}{N} \sum_{i=1}^N y_i \cdot \log(p(x_i) + (1 - y_i) \cdot \log(1 - p(x_i))
\end{equation}

During inference, the predicted probability values are thresholded to create the final binary segmentation.

\subsection{Categorical segmentation}
\label{sec:categorical_segmentation}
In the second experiment a semantic or categorical segmentation is applied to be able to separate the individual instances. Based on the direction-based classes described in Uhrig et al. \cite{uhrig_pixel-level_2016}, the semantic segmentation is set up with the classes \emph{background}, \emph{outer edge of an animal} and \emph{inner core of an animal} (see Figure \ref{fig:intro_ellipses_segmentation_modes_c}) to recognize the outer boundaries of the animals. Or in other words, it defines a distance-based classification which encodes the distance to the pigs center in discrete steps. Whereby the inner-core area is just a scaled down version of the original manually annotated ellipse. With these three classes, the training data are categorical label-images with an one-hot vector $t_i$ at each pixel, indicating one positive class and two negative classes.

In the existing network architecture, only the last layer is adapted, such that the number of channels corresponds to the number of classes $(C = 3)$ defined in the experiment. Since each pixel can only belong to one of the $C$ classes, the vector $x_i$ along the channel axis at each pixel-location is interpreted as a probability distribution over the $C$ classes. Such a probability distribution can be generated with the softmax activation function on the output-layer. The loss function is the categorical cross-entropy loss over all $N$ pixels and the $C$ classes:
\begin{equation}
L = -\frac{1}{N} \sum_{i=1}^N \sum_{j=1}^C t_{i, j} \cdot \log(x_{i, j})
\end{equation}
While in the binary segmentation the individual instances blend when they overlap, the centers of the animals and thus the individual instances can still be reconstructed with this method. A detailed description of the extraction process follows in subsection \ref{sec:ellipse_extraction}.
\subsection{Instance segmentation}
\label{sec:instance_segmentation_with_discriminative_loss}
\begin{figure}[ht]
\centering
\includegraphics[width=10cm]{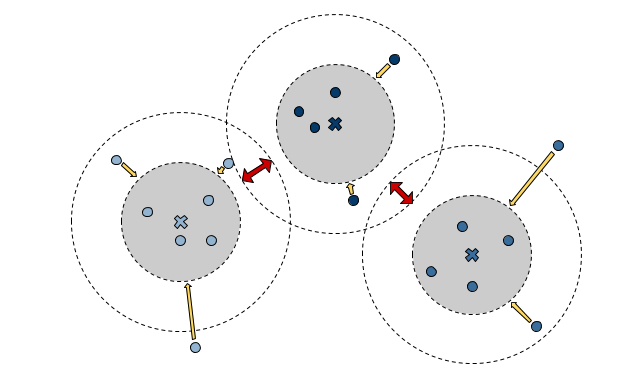}
\caption{\label{fig:discriminative}Illustration of the forces acting on the pixels to form the clusters (image adopted from \cite{de_brabandere_semantic_2017}).  With the variance term (yellow arrows) the pixels are drawn in the direction of the cluster mean (crosses). The distance term (red arrows) pushes the different clusters apart. Both forces are only active as long as the threshold values are not reached (inner circle for the cluster variance and outer circle for the distance).}
\end{figure}

The categorical segmentation is a rather naive approach, where the boundaries should prevent the individual animals from blending together. So in the third experiment each pixel in the image should be assigned to a specific animal (or the background). For this task De Brabandere et al. \cite{de_brabandere_semantic_2017} have introduced a discriminating loss function which uses a high dimensional feature space in which the pixel of the input image are projected in (pixel embedding). The network learns to place the pixels belonging to one object in this space as close together as possible, while pixels belonging to other objects are placed as far away as possible (see Figure \ref{fig:discriminative}).
\\
The loss function is a weighted combination of three terms, which act based on the individual instances given by the annotated data:
\begin{enumerate}[leftmargin=*,labelsep=4.9mm]
    \item \textbf{Variance term} The variance term penalizes the spatial variance of the pixel embeddings belonging to the same instance. For all pixels that belong to the object (according to the annotated data), the mean is calculated and then its distance of all object-pixels is evaluated. This forces the points in the feature space to cluster.
    \item \textbf{Distance term} The distance term keeps the calculated means of the clusters at a distance.
    \item \textbf{Regularization term} The regularization term keeps the expansion of all points in the feature space within limits and prevents them from drifting apart.
\end{enumerate}
Following the definition from \cite{de_brabandere_semantic_2017}, for each training example there are $C$ objects (or classes) to segment (the pigs plus the background). $N_c$ is the number of pixels, covering object $c$ and $x_i$ is one pixel embedding in the feature space. For each object $c$ there is a mean of all its pixel embeddings $\mu_c$. $\lVert \cdot \rVert$ is the L1 norm. In addition the loss is hinged to be less constrained in the representation. The pixel embeddings of the objects do not need to converge to exactly one point but should reach a distance below a threshold $\delta_v$. In the same way, the distance between two different mean embeddings must only be greater than or equal to the threshold $\delta_d$. This is mapped with the hinge-function $[x]_+ = max(0, x)$. Now the three terms can be formally defined as follows:
\begin{equation}
    L_{reg} = \frac{1}{C} \sum_{c=1}^C \lVert \mu_c \rVert
\end{equation}
\\
\begin{equation}
    L_{var} = \frac{1}{C} \sum_{c=1}^C \frac{1}{N_c} \sum_{i=1}^{N_c}[\lVert \mu_c - x_i \rVert - \delta_v]^2_+
    \label{eq:discrimitive_loss_var}
\end{equation}
\\
\begin{equation}
    L_{dist} = \frac{1}{C(C - 1)} \mathop{\sum_{c_A=1}^C \sum_{c_B=1}^C}_{c_A \neq c_B} [2\delta_d - \lVert \mu_{c_A} - \mu_{c_B} \rVert]^2_+
    \label{eq:discrimitive_loss_dist}
\end{equation}
The final loss function $L$ with weights $\alpha, \beta$ and $\gamma$ is given as:
\begin{equation}
    L = \alpha \cdot L_{var} + \beta \cdot L_{dist} + \gamma \cdot L_{reg}
    \label{eq:discrimitive_loss}
\end{equation}

\paragraph{Postprocessing}
After the network has been used to create the pixel embedding on an input image, the individual instances must be extracted from it. De Brabandere et al. \cite{de_brabandere_semantic_2017} propose the use of the mean-shift algorithm to identify cluster-centers and afterwards assign all pixels belonging to the cluster (in terms of the $\delta_v$ threshold) to the same object. 
In this work the hierarchical clustering algorithm HDBSCAN \cite{hutchison_density-based_2013} is used instead, as it shows improved performance in high dimensional embedding spaces. HDBSCAN is a density based hierarchical clustering and therefore optimally suited for the required clustering. It starts with a thinning of the non-dense areas. Then the dense areas are linked to a tree, which is converted into a hierarchy of linked components. Thus, a condensed cluster tree can be created by the parameter of minimum cluster size, and from this tree the final flat clusters can be extracted.

\begin{figure}[t]
\begin{subfigure}[t]{0.163\textwidth}
\includegraphics[width=\textwidth]{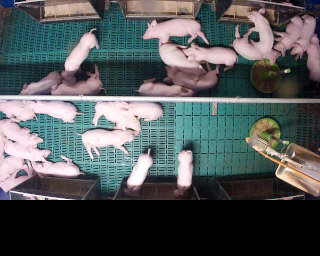}

\vspace{.1cm}
\includegraphics[width=\textwidth]{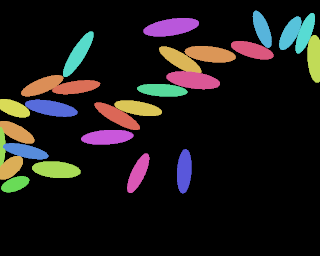}
\end{subfigure}
\begin{subfigure}[t]{0.163\textwidth}
\includegraphics[width=\textwidth]{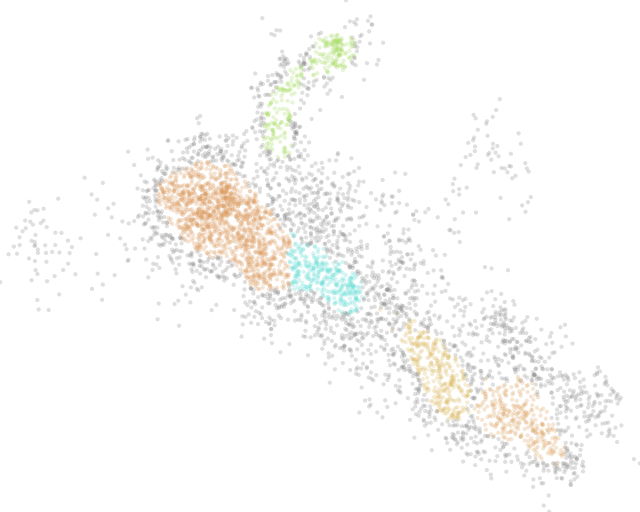}

\vspace{.1cm}
\includegraphics[width=\textwidth]{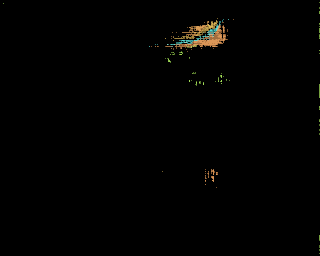}
\end{subfigure}
\begin{subfigure}[t]{0.163\textwidth}
\includegraphics[width=\textwidth]{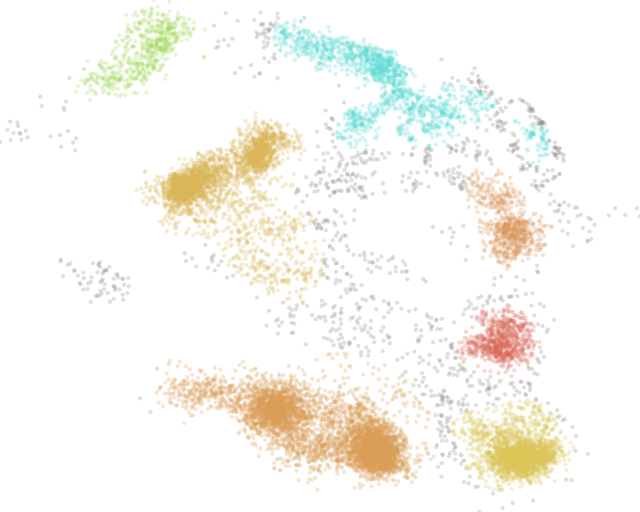}

\vspace{.1cm}
\includegraphics[width=\textwidth]{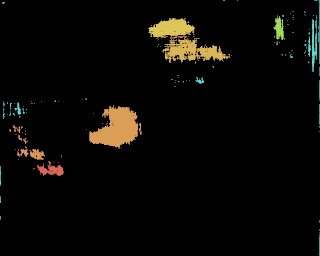}
\end{subfigure}
\begin{subfigure}[t]{0.163\textwidth}
\includegraphics[width=\textwidth]{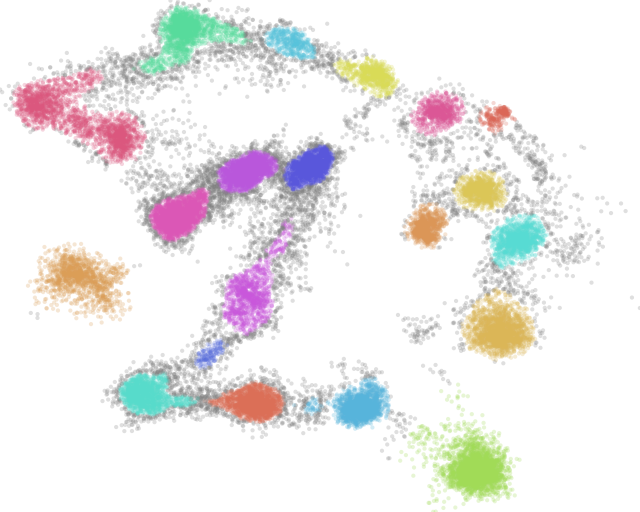}

\vspace{.1cm}
\includegraphics[width=\textwidth]{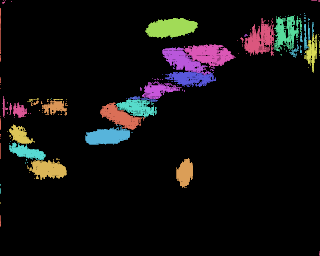}
\end{subfigure}
\begin{subfigure}[t]{0.163\textwidth}
\includegraphics[width=\textwidth]{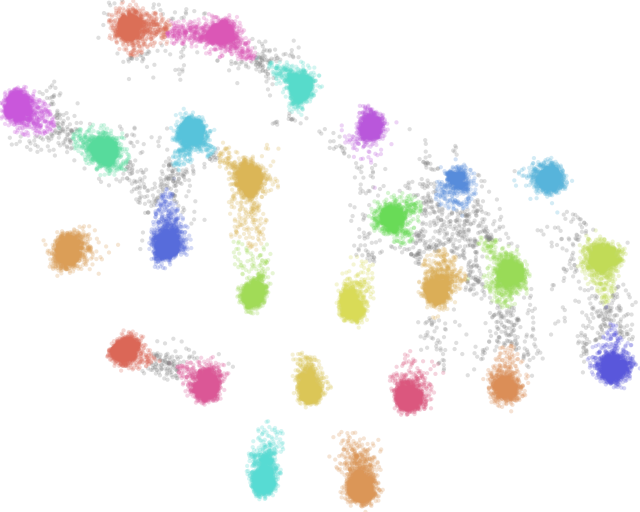}

\vspace{.1cm}
\includegraphics[width=\textwidth]{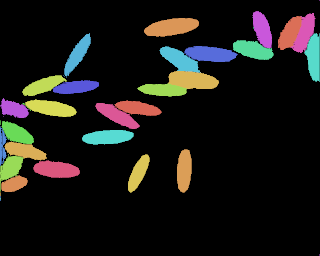}
\end{subfigure}
\begin{subfigure}[t]{0.163\textwidth}
\includegraphics[width=\textwidth]{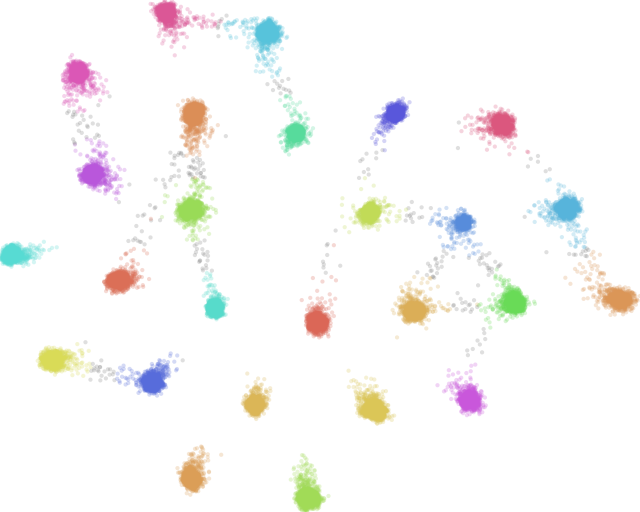}

\vspace{.1cm}
\includegraphics[width=\textwidth]{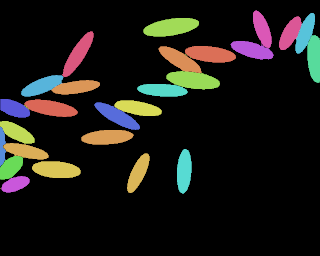}
\end{subfigure}
\caption{\label{fig:postprecessing_illustration}Results of the combined segmentation. On the left the original input image and below the ground truth label are shown. The top row depicts a two-dimensional embedding space. The bottom row depicts the corresponding binary segmentation and the assignment of the clusters. The snapshots are created after 1,2,3,10 and 80 gradient updates. The network was trained solely on the shown input image to generate the results shown here for illustration purposes.}
\end{figure}

\paragraph{Combined segmentation}
Since each pixel is mapped in the embedding, there are many data points that have to be clustered. At normal HD camera resolutions, this quickly adds up to a million data points. To accelerate clustering, a combined solution of discriminating and binary segmentation was designed. With the binary segmentation, a mask is created that contains only the pixels that belong to the animals. Thus only those pixels are fed into the clustering process that are relevant for the differentiation of the individual animals. Figure \ref{fig:postprecessing_illustration} shows an example of the distribution of pixels in a two-dimensional embedding and the clustering applied to the binary segmentation.

The network architecture only needs to be adapted slightly, since the architectures of the two experiments only differ in the last layer. In order to generate both outputs simultaneously, the network is equipped with two heads, which generate the corresponding outputs from the outputs of the autoencoder. The two heads are trained with the appropriate loss functions and feed the gradient updates equally weighted into the auto-encoder network.

\subsection{Orientation recognition}
If the found pixel segmentation approximately equals an ellipse shape, the fit of the final ellipses will align the ellipses so that the major axis corresponds to the orientation of the animal. However, since ellipses are symmetrical from a rotation of 180 degrees, the orientation of the animals can only be detected correctly up to this 180 degree ambiguity. Since the correct orientation was captured during manual annotation, this ambiguity can also be resolved. For this the combined method described in the previous section uses a categorical segmentation with the classes \emph{background}, \emph{body} and \emph{head} instead of a binary segmentation (see Figure \ref{fig:intro_ellipses_segmentation_modes_e}). In the postprocessing the classes then can be used to determine the orientation of the animals as described in subsection \ref{sec:ellipse_extraction}.

\section{Experimental Results}
\label{sec:experimental_results}
\subsection{Dataset}
The data used in this work are images from a conventional piglet rearing house. Five cameras were installed, with each camera covering two \SI{5.69}{\metre\squared} pens, each with a maximum of 13 animals. The animals were housed at the age of 27 days and remained in the facility for 40 days. The recordings of this dataset covered a period of four months. From all available videos 1000 frames with a resolution of 1280x800 pixels were randomly selected and manually annotated.  The images from one of the five cameras were declared as a test set, so that the evaluation is based on images of pens that the network never saw during the training. The images of the remaining four cameras make up the training and validation set. The data sets contain normal color images from the daytime periods and night vision images with active infrared illumination from the night periods. In addition, the cameras occasionally switched to night vision mode during the day due to dirty sensors. In the evaluation, however, a distinction is only made between color images and active night vision, regardless of the time of day. An overview can be found in Table \ref{tab:dataset_statistics}.

\begin{table}[h]
\caption{Dataset statistic for the 1000 randomly selected and annotated images. The images of the test set are taken from a different camera than the images of the training and validation set.}
\label{tab:dataset_statistics}
\vskip 0.15in
\begin{center}
\begin{small}
\begin{sc}
\begin{tabular}{lccc}
\toprule
Data set & total & daylight & nightvision \\
\midrule
train & 606 & 361 & 245 \\
validation & 168 & 96 & 72 \\
test & 226 & 108 & 118 \\
\bottomrule
\end{tabular}
\end{sc}
\end{small}
\end{center}
\vskip -0.1in
\end{table}

\begin{figure}[t]
\centering
\begin{subfigure}[t]{0.23\textwidth}
\includegraphics[width=\textwidth]{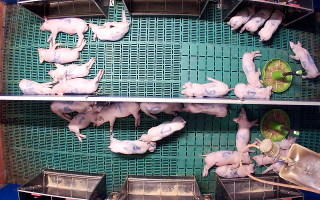}

\vspace{.1cm}
\includegraphics[width=\textwidth]{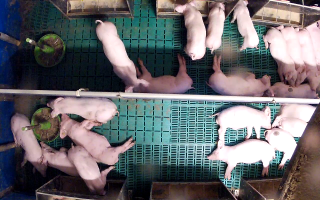}
\end{subfigure}
\hspace{.1cm}
\begin{subfigure}[t]{0.23\textwidth}
\includegraphics[width=\textwidth]{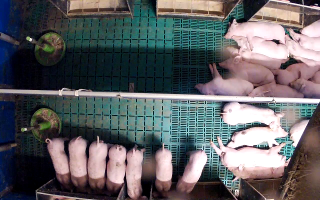}

\vspace{.1cm}
\includegraphics[width=\textwidth]{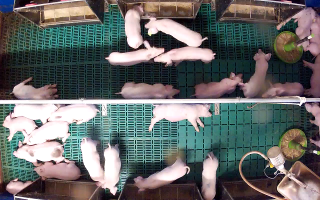}
\end{subfigure}
\hspace{.01cm}
\begin{subfigure}[t]{0.23\textwidth}
\includegraphics[width=\textwidth]{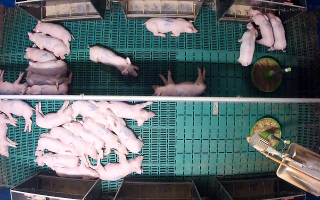}

\vspace{.1cm}
\includegraphics[width=\textwidth]{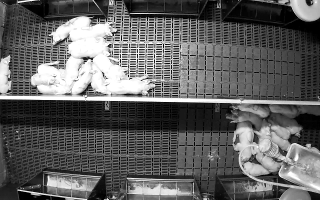}
\end{subfigure}
\hspace{.1cm}
\begin{subfigure}[t]{0.23\textwidth}
\includegraphics[width=\textwidth]{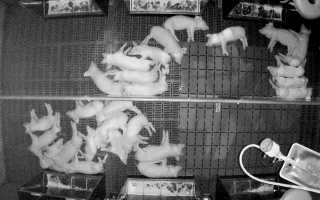}

\vspace{.1cm}
\includegraphics[width=\textwidth]{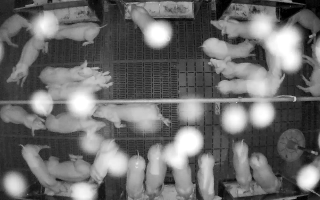}
\end{subfigure}
\caption{\label{fig:dataset_examples}Some sample images from the used data set. Note the poor lighting conditions, the dense grouping of the animals and the distortions during active night vision caused by dirt on the lens.}
\end{figure}

In Figure \ref{fig:dataset_examples} some example images from the data set are shown. Some of the challenges of working in pigsties can be clearly seen. For one thing, the camera position cannot always be chosen optimally, so that occlusions cannot be avoided. Furthermore, the lighting and the natural incidence of light cannot be controlled, so that the exposure conditions are sometimes difficult. And last but not least, the cameras get dirty over time, resulting in disturbances and malfunctions (such as, for example, the erroneously active night vision).

\subsection{Evaluation metrics}
\label{sec:evaluation_metrics}
For the task of \emph{panoptic segmentation} Kirillov et al. \cite{kirillov_panoptic_2019} also proposed a metric called \emph{panoptic quality} (PQ). It is very similar to the well known F1 Score, but takes into account the special characteristic that each pixel can only be assigned to exactly one object. It first matches the predicted segments with the ground truth segments and afterwards calculates a score based on the matches. 

Since each pixel can only be assigned to one object, the predicted segments cannot overlap. Therefore it can be shown that there can be at most one predicted segment for each ground truth segment, with an intersection over union (IoU) of strictly greater than 0.5 \cite{kirillov_panoptic_2019}. Each ground truth segment for which there is such a matching predicted segment counts as a \emph{true positive} (TP). Predicted segments that do not sufficiently overlap any ground truth segment count as \emph{false positives} (FP) and uncovered ground truth segments count as \emph{false negatives} (FN). For all the predicted segments $p$ and the ground truth segments $g$, the PQ is defined as:
\begin{equation}
    PQ = \frac{\sum_{(p,g) \in TP} IoU(p,g)}{|TP| + \frac{1}{2}|FP| + \frac{1}{2}|FN|}
\end{equation}
For better comparability with other work, the F1 Score, precision and recall are also evaluated in the experiments (see Subsection \ref{sec:evaluation}). F1, precision and recall are based on the same TP, FP, and FN as the PQ.

\subsection{Ellipse extraction}
\label{sec:ellipse_extraction}
\begin{figure}[ht]
\centering
\begin{subfigure}[t]{0.19\textwidth}
\includegraphics[width=\textwidth]{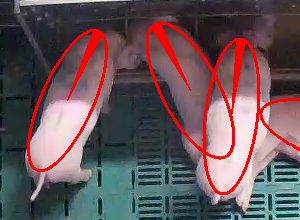}

\vspace{.1cm}
\includegraphics[width=\textwidth]{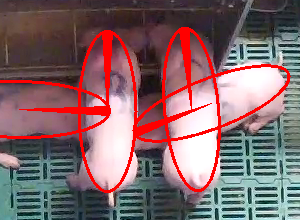}
\caption{\label{fig:ellipse_extraction_manual_annotated_ellipses}Manual annotated ellipses}
\end{subfigure}
\hspace{.01cm}
\begin{subfigure}[t]{0.19\textwidth}
\includegraphics[width=\textwidth]{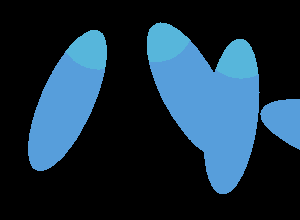}

\vspace{.1cm}
\includegraphics[width=\textwidth]{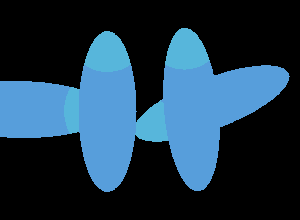}
\caption{\label{fig:ellipse_extraction_orientation_recognition}Label images for orientation recognition}
\end{subfigure}
\hspace{.01cm}
\begin{subfigure}[t]{0.19\textwidth}
\includegraphics[width=\textwidth]{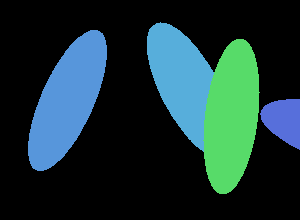}

\vspace{.1cm}
\includegraphics[width=\textwidth]{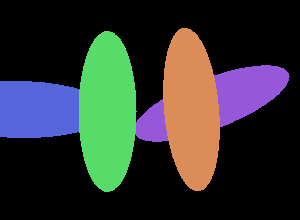}
\caption{\label{fig:ellipse_extraction_instance_labels}Label images for instance segmentation}
\end{subfigure}
\hspace{.01cm}
\begin{subfigure}[t]{0.19\textwidth}
\includegraphics[width=\textwidth]{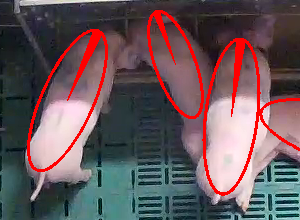}

\vspace{.1cm}
\includegraphics[width=\textwidth]{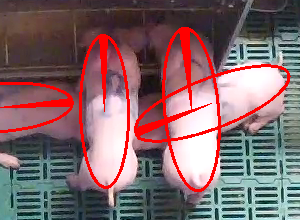}
\caption{\label{fig:ellipse_extraction_extracted_gt_ellipses}Extracted ground truth ellipses}
\end{subfigure}
\hspace{.01cm}
\begin{subfigure}[t]{0.19\textwidth}
\includegraphics[width=\textwidth]{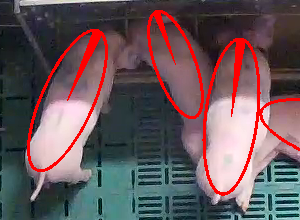}

\vspace{.1cm}
\includegraphics[width=\textwidth]{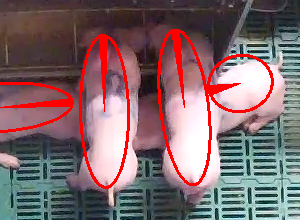}
\caption{Predicted ellipses by the proposed method}
\end{subfigure}
\caption{\label{fig:ellipse_extraction}Two examples of the manual annotated ellipses (a), the created label images for orientation recognition (b), instance segmentation (c), the extracted ground truth ellipses (d) and the results (e). The filled part of the ellipses shows the identified orientation of the animals. Note the adjusted overlaps in (d), which allow a comparison with the predicted ellipses. The lower picture in (e) shows a faulty detection.}
\end{figure}
To capture all pixels belonging to an animal in the manual annotation, the ellipses must be able to overlap (see Figure \ref{fig:ellipse_extraction_manual_annotated_ellipses}). While the depth sorting described in subsection \ref{sec:representation_of_the_pigs} ensures that each pixel is uniquely assigned to a single animal (see Figure \ref{fig:ellipse_extraction_instance_labels}), the pixel-level segmentations can not be compared to the originally annotated ground truth ellipses. If the animals overlap, the original ellipses and the found segmentations differ in size. To solve this issue and to generate comparable data, new ellipses were extracted from the label images by fitting ellipses into the segmentations (see Figures \ref{fig:ellipse_extraction_instance_labels} and \ref{fig:ellipse_extraction_extracted_gt_ellipses}). These new ground truth ellipses are then compared to the ellipses extracted from the segmentation-output of the networks.

Depending on the experiment the ellipses are extracted differently from the predicted outputs of the network. For the categorical segmentation all pixels of the class \emph{inner core of an animal} (see section \ref{sec:categorical_segmentation}) are searched first using a blob search. The individual separate blobs are then interpreted as individual animals. For this an ellipse is fitted to the segmented pixel with the algorithm of Fitzgibbon \cite{fitzgibbon_buyers_1996}. Since the core of an animal was generated from the scaled-down version of the manually annotated ellipse, the ellipse adapted from the blob can then simply be scaled up accordingly. 

When using the segmentation with the discriminative loss and the clustering, the ellipses can simply be fitted to the pixels of the individual clusters, after backprojecting the pixels from the embedding into image-space. As described in subsection \ref{sec:instance_segmentation_with_discriminative_loss}, the binary mask of the combined approach is used thereby to process only the pixels that belong to the animals while masking out the background. If the orientation of the animals is also detected, the classes \emph{body} and \emph{head} can be combined to achieve the binary segmentation. Once the ellipses are fitted, the original categorical segmentation can be used to identify the side of the ellipse where the head was detected.

\subsection{Implementation Details}
\label{sec:implementation_details}

The network was implemented with the \emph{segmentation models} library \cite{yakubovskiy_segmentation_2019}. As described in subsection \ref{sec:network_architecture} an U-Net with ResNet34 and Inception-ResNet-v2 encoder-backbones was used. The backbones were initialized with weights pretrained on ImageNet \cite{deng_imagenet_2009}. Four skip connections were added for both backbones, one after each major resolution reduction. In the case of ResNet34 accordingly after each of the four stages \cite{he_deep_2015}. For Inception-ResNet-v2 two directly in the \emph{Stem}-block, one after the ten repetitions of the \emph{Inception-A} block and one after the 20 repetitions of the \emph{Inception-B} block \cite{szegedy_inception-v4_2016}. Exact details on the structure of the blocks in the encoder backbones can be found in the corresponding papers. The decoders are assembled from similar blocks, but instead of MaxPooling layers, they use upSampling layers between blocks to reproduce the original resolution. For all experiments the \emph{Adam}-Optimizer \cite{kingma_adam_2017} with an initial learning rate of 1e-4 was used.

To speed up the calculation of the network and any subsequent clustering, the images were scaled down to a resolution of 640 x 512 pixels. Additionally the training images were augmented during the training with the \emph{imgaug} library \cite{jung_imgaug_2020}, to achieve a better generalization. The augmentation included different distortions, affine transformations and color changes (e.g. grayscale to simulate active infrared illumination) and increased the amount of training images by a factor of 10. For all the image related pre- and post-processing tasks (such as the ellipse fitting) the OpenCV-library \cite{bradski_opencv_2000} was used. 

For the pixel embedding an eight-dimensional space was used. The thresholds in the discriminative loss in Equation \ref{eq:discrimitive_loss_var} and \ref{eq:discrimitive_loss_dist} were set to $\delta_v = 0.1$ and $\delta_d = 1.5$. The weights in the final loss term in equation \ref{eq:discrimitive_loss} were set to $\alpha = \beta = 1.0$ and $\gamma = 0.001$. The values were taken from the original paper \cite{de_brabandere_semantic_2017}, except for the threshold $\delta_v$, which was decreased to improve the density-based clustering. For the clustering the HDBSCAN implementation from McInnes et al. \cite{mcinnes_hdbscan_2017} was used with the minimal cluster size set to $100$.

\subsection{Evaluation}
\label{sec:evaluation}
In order to evaluate the methods described in Chapter \ref{sec:proposed_method}, they were all run on the test dataset. To investigate the influence of different backbones, all experiments were performed with both backbones. A distinction was also made between day and night vision images to test the robustness of the methods.
\paragraph{Binary segmentation}
In binary segmentation, the network predicts a probability that a particular pixel belongs to a pig or the background. This probability is converted into a final decision using a threshold value of $0.5$. The binary pixel values can then be compared with the ground-truth images using the Jaccard index. This gives the accuracy of the predictions as listed in Table \ref{tab:evaluation_binary_segmentation}.

\begin{table}[h]
\caption{Accuracy-results of the binary segmentation experiment (measured with the Jaccard index). The experiment was carried out on all test images, and separately on the daylight (D) and night vision (N) images only.}
\label{tab:evaluation_binary_segmentation}
\vskip 0.15in
\begin{center}
\begin{small}
\begin{sc}
\begin{tabular}{lccc}
\toprule
Backbone & Acc & Acc (d) & Acc (n) \\
\midrule
ResNet34 & 0.9730 & 0.9771 & 0.9692 \\
Incep.-RN-v2 & \textbf{0.9735} & \textbf{0.9774} & \textbf{0.9699} \\
\bottomrule
\end{tabular}
\end{sc}
\end{small}
\end{center}
\vskip -0.1in
\end{table}

\begin{figure}[t]
\centering
\begin{subfigure}[t]{0.19\textwidth}
\includegraphics[width=\textwidth]{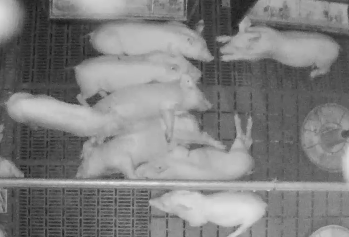}
\caption{\label{fig:evaluation_examples_img}Original image part}
\end{subfigure}
\hspace{.01cm}
\begin{subfigure}[t]{0.19\textwidth}
\includegraphics[width=\textwidth]{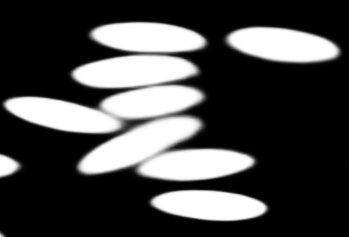}
\caption{\label{fig:evaluation_examples_binary}Binary segmentation}
\end{subfigure}
\hspace{.01cm}
\begin{subfigure}[t]{0.19\textwidth}
\includegraphics[width=\textwidth]{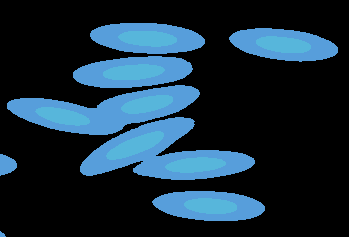}
\caption{\label{fig:evaluation_examples_categorical}Categorical segmentation}
\end{subfigure}
\hspace{.01cm}
\begin{subfigure}[t]{0.19\textwidth}
\includegraphics[width=\textwidth]{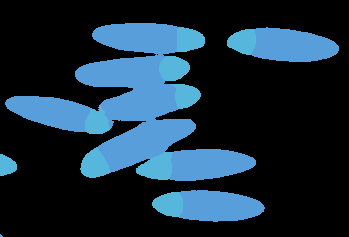}
\caption{\label{fig:evaluation_examples_withhead}Orientation recognition}
\end{subfigure}
\hspace{.01cm}
\begin{subfigure}[t]{0.19\textwidth}
\includegraphics[width=\textwidth]{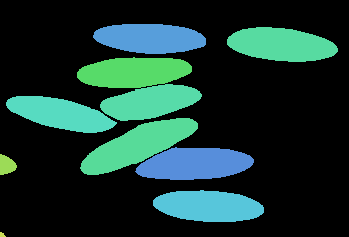}
\caption{\label{fig:evaluation_examples_instance}Combined segmentation}
\end{subfigure}
\caption{\label{fig:evaluation_examples}Results from the different experiments on an example image (cropped) (a). Depicted are the simple binary segmentation (b), the categorical segmentation with classes \emph{outer edge of an animal} and \emph{inner core of an animal} (c), the body part segmentation for the orientation recognition with classes \emph{head} and \emph{rest of the body} (d) and the combined segmentation with the results of the clustering, masked with the binary segmentation (d).}
\end{figure}

The ellipses cover the body of the animals only approximately (see subsection \ref{sec:representation_of_the_pigs}). Therefore, the network sometimes receives ambiguous information, where pixels that can be clearly recognized as background still have the label \emph{pig}. The network produces mainly elliptical predictions, but the segmented areas also follow the body of the animals (see Figure \ref{fig:evaluation_examples_binary}). Since the label images only contain undistorted ellipses, an accuracy of 100\% is never achievable for the network.

\paragraph{Categorical segmentation}

For the categorical segmentation the class \emph{inner core of an animal} was set to 50\% of the size of the ellipses (see Figure \ref{fig:evaluation_examples_categorical}). The results are shown in the upper part of Table \ref{tab:evaluation_categorical_combined_segmentation}. Beside the accuracy of the categorical segmentation (again measured with the Jaccard index), now also the extracted ellipses (see subsection \ref{sec:ellipse_extraction}) were compared to the manual annotated ellipses with the \emph{panoptic quality} metric. F1 score, precision and recall are listed in detail in Table \ref{tab:evaluation_categorical_combined_detection}.

\begin{table}[h]
\caption{Detection results for the ellipses extracted with the categorical segmentation and the combined segmentation. Regardless of the selected backbone, detection rates of about 95\% (F1 Score) are achieved. For detailed information about precision and recall see Table \ref{tab:evaluation_categorical_combined_detection}. It is noticeable that with the combined segmentation approach the accuracy of the binary segmentation remains unaffected, although the segmentation head and the pixel-embedding head jointly influence the weights in the backbone. The experiments were carried out on all test images, and separately on the daylight (D) and night vision (N) images only.}
\label{tab:evaluation_categorical_combined_segmentation}
\vskip 0.15in
\centering
\tablesize{\footnotesize}
\begin{tabular}{lccccccccc}
\toprule
Categorical & PQ &  PQ (d) & PQ (n) & F1 &  F1 (d) & F1 (n) & Cat. Acc & Cat. Acc (d) & Cat. Acc (n) \\
\midrule
ResNet34 & 0.7920 & 0.8124 & 0.7738 & \textbf{0.9550} & \textbf{0.9619} & \textbf{0.9487} & 0.9612 & 0.9664 & 0.9565 \\
Incep.-RN-v2 & 0.7943 & 0.8165 & 0.7742 & 0.9541 & 0.9614 &  0.9475 & 0.9612 & 0.9664 & 0.9564 \\
\midrule
Combined & PQ &  PQ (d) & PQ (n) & F1 &  F1 (d) & F1 (n) & Bin. Acc & Bin. Acc (d) & Bin. Acc (n) \\
\midrule
ResNet34 & \textbf{0.7966} & \textbf{0.8181} & \textbf{0.7774} & 0.9513 & 0.9588 & 0.9446 & \textbf{0.9722} & \textbf{0.9761} & \textbf{0.9687} \\
Incep.-RN-v2 & 0.7921 & 0.8179 & 0.7689 & 0.9481 & 0.9566 & 0.9404 & 0.9707 & 0.9752 & 0.9666 \\
\bottomrule
\end{tabular}
\vskip -0.1in
\end{table}

\paragraph{Instance segmentation}
For this experiment a combined network was trained to predict the association of each pixel with the individual animals in an eight-dimensional space together with the binary segmentation. The results are shown in the lower part of Table \ref{tab:evaluation_categorical_combined_segmentation}. F1 score, precision and recall are listed in the lower part of Table \ref{tab:evaluation_categorical_combined_detection}. It is important to note that the combined processing of pixel embedding and binary segmentation in a shared backbone does not affect the accuracy of the binary segmentation. Therefore, a synergy effect of the two tasks can be assumed.

\paragraph{Orientation recognition}
For the orientation recognition, the same combined network as before was used, but the binary segmentation was replaced by the body part segmentation (see Figure \ref{fig:evaluation_examples_withhead}). The orientation of the ellipses is the reconstructed as described in subsection \ref{sec:ellipse_extraction}. To evaluate the accuracy of the orientation recognition, the orientation of all correctly identified pigs (true positives) was assessed over the complete test-set. The results are summarized in Table \ref{tab:evaluation_withhead_segmentation}. Although a categorical segmentation is now applied instead of the binary segmentation, a comparison with the values in Table \ref{tab:evaluation_categorical_combined_segmentation} shows that the accuracy of ellipse detection is not affected.

\begin{table}[h]
\caption{Results of orientation recognition. The network can correctly recognize orientation in 94\% of the correctly found animals (true positive).}
\label{tab:evaluation_withhead_segmentation}
\vskip 0.15in
\begin{center}
\begin{small}
\begin{sc}
\begin{tabular}{lccc}
\toprule
Backbone & Orien. Acc & PQ & Cat. Acc\\
\midrule
ResNet34 & \textbf{0.9428} & \textbf{0.7958} & \textbf{0.9644} \\
Incep.-RN-v2 & 0.9226 & 0.7898 & 0.9601 \\
\bottomrule
\end{tabular}
\end{sc}
\end{small}
\end{center}
\vskip -0.1in
\end{table}

\subsection{Ablation studies}
\label{sec:ablation_studies}
The experiments conducted with the two differently complex encoder architectures already suggest that the influence of the backbone is marginal. Nevertheless, additional experiments were carried out to confirm this assumption. To increase the speed of the tests, the resolution of the input images was further reduced to 320x256 pixels. The results are summarized in Table \ref{tab:ablation_studies}.
\paragraph{Classification backbone} As described in subsection \ref{sec:network_architecture} the chosen U-Net architecture can be set up with different classification backbones. In addition to the classification backbones already introduced, the experiments were also carried out with the EfficientNet \cite{tan_efficientnet_2019} backbone.
\paragraph{Network architecture} Although the U-Net architecture delivers good results, all three backbones were additionally evaluated with the FPN architecture (see subsection \ref{sec:network_architecture}).

\paragraph{Clustering hyperparameters} To optimize the density-based clustering, the thresholds $\delta_v$ and $\delta_d$ in the discriminative loss are available as hyperparameters (see subsection \ref{sec:instance_segmentation_with_discriminative_loss}). They control how close the clusters are moved together, or how much distance different clusters have to keep from each other. These two threshold values were also evaluated in a grid search with the result that the exact values have no influence on the accuracy of the clustering. Values between $0.05$ and $0.3$ for $\delta_v$ and values between $1$ and $3$ for $\delta_d$ all yielded approximately the same result.

There is also the minimal cluster size, which refers to the number of pixels that at least belong to one pig. This parameter is therefore primarily dependent on the resolution of the input images and can only be set to a limited extent as a hyperparameter.
\begin{table}[h]
\caption{Results of the ablation study with the combined segmentation on the test dataset. For this evaluation a reduced image resolution of 320x256 pixels was used. The results highlight the marginal impact of the different architecture choices (U-Net vs. FPN) as well as the different backbones.}
\label{tab:ablation_studies}
\vskip 0.15in
\begin{center}
\begin{small}
\begin{sc}
\begin{tabular}{lccccc}
\toprule
Combined U-Net & PQ & F1 & Precision & Recall & Cat. Acc \\
\midrule
ResNet34 & 0.7863 & 0.9457 & 0.9559 & 0.9358 & 0.9694 \\
Incep.-RN-v2 & 0.7685 & 0.9326 & 0.9501 & 0.9157 & 0.9674 \\
EfficientNet-B5 & 0.7768 & 0.9404 & 0.9471 & 0.9337 & 0.9692 \\
\midrule
Combined FPN & PQ & F1 & Precision & Recall & Cat. Acc \\
\midrule
ResNet34 & 0.7824 & 0.9442 & 0.9556 & 0.9332 & 0.9709 \\
Incep.-RN-v2 & 0.7784 & 0.9414 & 0.9511 & 0.9319 & 0.9700 \\
EfficientNet-B5 & 0.7861 & 0.9451 & 0.9556 & 0.9347 & 0.9709 \\
\bottomrule
\end{tabular}
\end{sc}
\end{small}
\end{center}
\vskip -0.1in
\end{table}

\section{Discussion}
\label{sec:discussion}
\begin{figure}[t]
\centering
\begin{subfigure}[t]{0.32\textwidth}
\includegraphics[width=\textwidth]{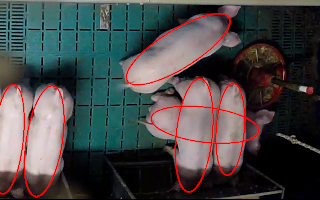}
\caption{Original image part with annotated ellipses}
\label{fig:overlap_problem_example_orig}
\end{subfigure}
\hspace{.01cm}
\begin{subfigure}[t]{0.32\textwidth}
\includegraphics[width=\textwidth]{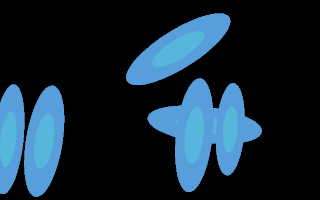}
\caption{Label image for the categorical segmentation}
\label{fig:overlap_problem_example_fcn}
\end{subfigure}
\hspace{.01cm}
\begin{subfigure}[t]{0.32\textwidth}
\includegraphics[width=\textwidth]{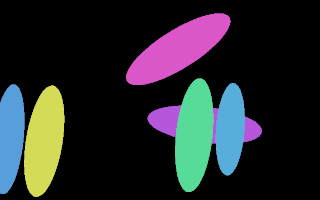}
\caption{Label image for instance segmentation}
\label{fig:overlap_problem_example_clusters}
\end{subfigure}
\caption{\label{fig:overlap_problem_example}Example of the fragility of the categorical segmentation in case of strong overlaps. If the center of the animals is not visible, the segmentation cannot provide meaningful information about the hidden animal (b). The instance segmentation, on the other hand, does not have this problem (c).}
\end{figure}

\begin{figure}[h]
\centering
\begin{subfigure}[t]{0.29\textwidth}
\includegraphics[width=\textwidth]{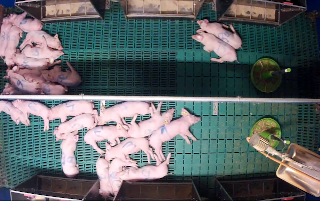}

\vspace{.1cm}
\includegraphics[width=\textwidth]{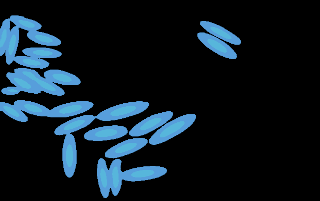}

\vspace{.1cm}
\includegraphics[width=\textwidth]{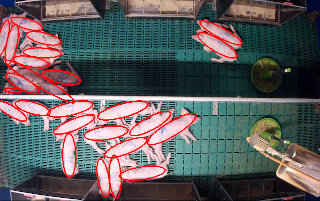}

\vspace{.1cm}
\includegraphics[width=\textwidth]{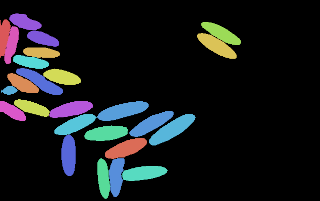}

\vspace{.1cm}
\includegraphics[width=\textwidth]{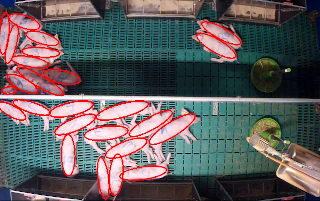}
\caption{\label{fig:result_examples_1}Example of an error caused by the blending of two animals in the top left group.}
\end{subfigure}
\hspace{.01cm}
\begin{subfigure}[t]{0.29\textwidth}
\includegraphics[width=\textwidth]{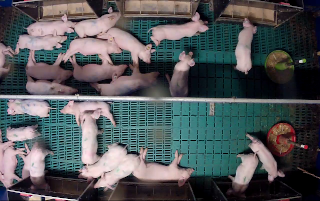}

\vspace{.1cm}
\includegraphics[width=\textwidth]{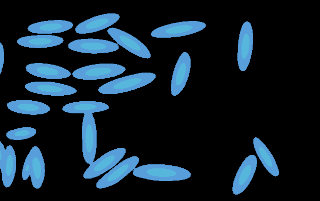}

\vspace{.1cm}
\includegraphics[width=\textwidth]{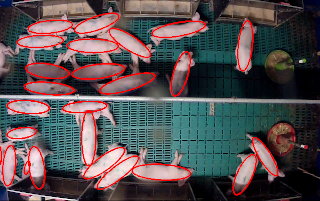}

\vspace{.1cm}
\includegraphics[width=\textwidth]{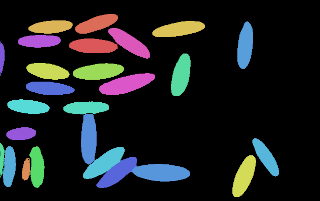}

\vspace{.1cm}
\includegraphics[width=\textwidth]{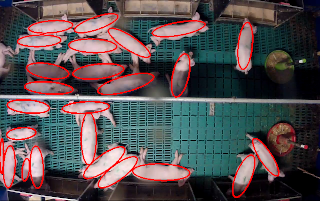}
\caption{\label{fig:result_examples_2}Example for animals that are lying at the edge of the field of view (left) and are therefore difficult to recognize.}
\end{subfigure}
\hspace{.001cm}
\begin{subfigure}[t]{0.29\textwidth}
\includegraphics[width=\textwidth]{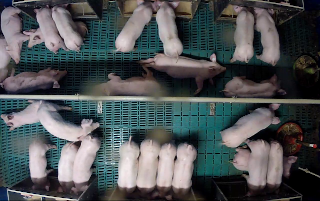}

\vspace{.1cm}
\includegraphics[width=\textwidth]{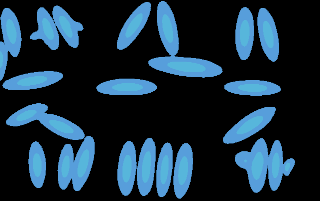}

\vspace{.1cm}
\includegraphics[width=\textwidth]{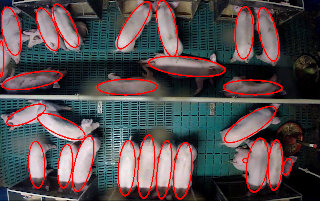}

\vspace{.1cm}
\includegraphics[width=\textwidth]{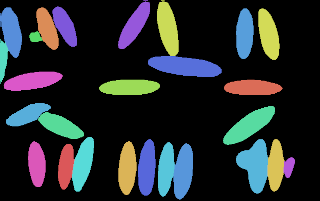}

\vspace{.1cm}
\includegraphics[width=\textwidth]{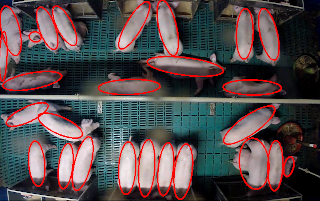}
\caption{\label{fig:result_examples_3}Example of complex overlaps (top left and bottom right).}
\end{subfigure}
\caption{\label{fig:result_examples}Examples of the difficulties that the data set contains. The pictures show from top to bottom: The original image, the prediction of the categorical segmentation, the ellipses extracted from the categorical segmentation, the prediction of the combined segmentation, the ellipses extracted from the combined segmentation.}
\end{figure}

As shown in Table \ref{tab:evaluation_categorical_combined_segmentation}, the quality of the extracted ellipses of the categorical segmentation and that of the combined approach are comparable on average. Especially for more complex overlaps, the categorical segmentation theoretically reaches its limits when the core part of the pigs is hardly visible (see Figure \ref{fig:overlap_problem_example_fcn}). In such situations the pixel embedding should have shown its strengths, but these situations hardly seem to occur in the actual data set. Therefore, the network was not able to learn these cases and produces correspondingly bad results (see Figure \ref{fig:result_examples_3}).

It is interesting too that the choice of the backbone and overall architecture also has no influence on the results (see ablation study in Table \ref{tab:ablation_studies}). The small size of the data set will probably play a role here. With so little data (although augmented), the deep architectures like Inception-ResNet-v2 and EfficientNet cannot even show their advantages over ResNet34.

The choice of the PQ as evaluation metric makes sense with the methods presented, since the exact evaluation of the \emph{Intersection over Union} provides information about how precisely the pixel-accurate segmentation works. Unfortunately, this novel metric does not allow a direct comparison to other works. However, in order to allow a rough comparison, classical metrics like Precision and Recall are listed in Table \ref{tab:evaluation_categorical_combined_detection}. In the only paper with a publicly accessible data set \cite{psota_multi-pig_2019}, the authors give 91\% precision and 67\% for their test set. With our methods on our data set we achieve values around 95\% for both metrics. However, it should be noted that although the test data in our data set comes from a different camera, the images in the test set do not differ fundamentally from the images in the training set. In \cite{psota_multi-pig_2019}, the images in the test set seem to deviate more from the training data.
A direct comparison on the data set from \cite{psota_multi-pig_2019} was unfortunately not possible because ellipses cannot be reconstructed easily from the given key points. For a comparison, their data set would have had to be annotated completely by hand with ellipses. Other public data sets of pigs do not exist to our knowledge.
\begin{table}[h]
\caption{Detailed listing of F1 Score, precision and recall. The experiment was carried out on all test images, and separately on the daylight (D) and night vision (N) images only.}
\label{tab:evaluation_categorical_combined_detection}
\vskip 0.15in
\centering
\tablesize{\footnotesize}
\begin{tabular}{lccccccccc}
\toprule
Categorical & F1 &  F1 (d) & F1 (n) & Prec & Prec (d) & Prec (n) & Recall & Recall (d) & Recall (n) \\
\midrule
ResNet34 & \textbf{0.9550} & \textbf{0.9619} & \textbf{0.9487} & \textbf{0.9586} & \textbf{0.9678} & 0.9503 & \textbf{0.9514} & 0.9560 & \textbf{0.9472}\\
Incep.-RN-v2 & 0.9541 & 0.9614 & 0.9475 & 0.9577 & 0.9626 & \textbf{0.9532} & 0.9505 & \textbf{0.9601} & 0.9418\\
\midrule
Combined & F1 &  F1 (d) & F1 (n) & Prec & Prec (d) & Prec (n) & Recall & Recall (d) & Recall (n) \\
\midrule
ResNet34 & 0.9513 & 0.9588 & 0.9446 & 0.9544 & 0.9645 & 0.9454 & 0.9482 & 0.9531 & 0.9438\\
Incep.-RN-v2  & 0.9481 & 0.9566 & 0.9404 & 0.9495 & 0.9598 & 0.9402 & 0.9466 & 0.9535 & 0.9405\\
\bottomrule
\end{tabular}
\vskip -0.1in
\end{table}

In general, the correct evaluation is difficult because there is no defined set of rules for annotation. In \cite{psota_multi-pig_2019}, for example, the pigs that are in the field of view but not in the observed bay were not annotated. A network that recognizes these pigs anyway would be punished with \emph{false positives} here. Furthermore, there are also borderline cases in our data set where pigs are hardly visible but still marked by the human annotators. If such pigs are not found due to sanity checks like a minimum pixel number in clustering or a bad segmentation, false negatives are counted (see Figure \ref{fig:result_examples_2}). Here, a publicly accessible data set with fixed rules would be useful in the future.

\section{Conclusions}
\label{sec:conclusions}
The methods shown here have achieved very good results on the data used and offer a pixel accurate segmentation of the animals instead of bounding boxes or keypoints. The already described advantage over the existing methods is that more information can be extracted from the segmentation. For example, conclusions can be drawn about the volume and thus the weight of the animals. Weight gain and other health factors can thus be determined and evaluated.

The ablation study has shown that all variants provide approximately the same results. From this it can be concluded that the data used so far do not provide more variance to learn the errors and inconsistencies that occur. An increase of the data set or an enrichment in variance would be the next step to check the generalization of the presented methods.

\vspace{6pt} 



\authorcontributions{conceptualization, J.B., M.G. and I.T.; methodology, J.B.; software, J.B.; data acquisition, M.G. and I.T.; data annotation, J.B. and M.G.; writing--original draft preparation, J.B.; writing--review and editing, J.B., M.G., I.T and R.K.; visualization, J.B.; supervision, R.K.}


\acknowledgments{We would like to thank Lars Schmarje and Monty Santarossa for their valuable feedback.}

\conflictsofinterest{The authors declare no conflict of interest.}




\externalbibliography{yes}
\bibliography{references}



\end{document}